\documentclass[conference]{IEEEtran}
\IEEEoverridecommandlockouts
\usepackage{cite}
\usepackage{amsmath,amssymb,amsfonts}
\usepackage{algorithmic}
\usepackage{graphicx}
\usepackage{textcomp}
\usepackage{xcolor}
\usepackage{bm, multirow}
\usepackage{booktabs,tabu}
\usepackage{arydshln}
\def\BibTeX{{\rm B\kern-.05em{\sc i\kern-.025em b}\kern-.08em
    T\kern-.1667em\lower.7ex\hbox{E}\kern-.125emX}}
\begin{document}

\title{Pseudo-label Learning with Calibrated Confidence \\Using an Energy-based Model
\thanks{This work was supported in part by JSPS KAKENHI Grant Numbers JP21H03511, JP21K18312, JP22H05172, and JP22H05173.}
}

% \author{\IEEEauthorblockN{Anonymous Authors}}

\author{\IEEEauthorblockN{Masahito Toba}
\IEEEauthorblockA{
% \textit{Department of Information Science and Technology} \\
\textit{Kyushu University}\\
Fukuoka, Japan \\
masahito.toba@human.ait.kyushu-u.ac.jp}
\and
\IEEEauthorblockN{Seiichi Uchida}
\IEEEauthorblockA{
% \textit{Department of Advanced Information Technology} \\
\textit{Kyushu University}\\
Fukuoka, Japan \\
uchida@ait.kyushu-u.ac.jp}
\and
\IEEEauthorblockN{Hideaki Hayashi}
\IEEEauthorblockA{
% \textit{Institute for Datability Science} \\
\textit{Osaka University}\\
Suita, Japan \\
hayashi@ids.osaka-u.ac.jp}
}
% \and
% \and
% \and

\maketitle

\begin{abstract}
In pseudo-labeling (PL), which is a type of semi-supervised learning, pseudo-labels are assigned based on the confidence scores provided by the classifier; therefore, accurate confidence is important for successful PL. In this study, we propose a PL algorithm based on an energy-based model (EBM), which is referred to as the energy-based PL (EBPL). In EBPL, a neural network-based classifier and an EBM are jointly trained by sharing their feature extraction parts. This approach enables the model to learn both the class decision boundary and input data distribution, enhancing confidence calibration during network training. The experimental results demonstrate that EBPL outperforms the existing PL method in semi-supervised image classification tasks, with superior confidence calibration error and recognition accuracy.
\end{abstract}


\section{Introduction}
\label{sec:intro}
Effective learning using only a few labeled data is required in the field of pattern recognition using neural networks (NN). This is particularly important in domains such as medical data, where annotation costs can be prohibitively high. To overcome this challenge, various machine-learning methods have been proposed to reduce annotation costs. Semi-supervised learning~\cite{zhu2005ssl} employs a small number of labeled data and a large number of unlabeled data, transfer learning~\cite{pan2010trans} leverages labeled data from other domains, and active learning~\cite{settles2009active} prioritizes labeling data that are beneficial for learning.

PL~\cite{lee2013pseudo} is one of the popular methods of semi-supervised learning. In PL, the predicted class labels for unlabeled data are provisionally regarded as correct class labels and assigned as pseudo-labels to improve and facilitate the classifier training. For successful PL, it is necessary to select the appropriate samples to which pseudo-labels are assigned as accurately as possible.
 
The simplest method for selecting target samples for PL is to refer to the confidence that represents the uncertainty of the classifier's output~\cite{NEURIPS2020_fixmatch}. In NNs, it is common to use the maximum value of the class posterior (i.e., $\max_c~p(c \mid \bm{x})$ for class $c$ given an input sample $\bm{x}$) obtained as the output of the softmax function in the final layer as the confidence. Since higher confidence values are expected to correspond to more accurate predictions, samples with higher confidence are prioritized for assigning pseudo-labels. It is desirable for the confidence to match the actual classification accuracy. For example, accuracy for samples with confidence scores of around $0.6$ is expected to be about $60\%$. Referring to appropriately calculated confidence during training reduces the number of incorrect pseudo-labels and achieves successful classifier training.

However, recent studies have highlighted that the confidence provided by a deep NN is not necessarily consistent with the actual classification accuracy~\cite{pmlr-guo17a-oncalib}, which causes a critical problem when assigning pseudo-labels based on confidence. The issue arises from the fact that the cross-entropy loss function employed in NN training becomes smaller as the confidence approaches one (i.e., as the posterior for a specific class becomes one and for other classes approaches zero). This makes the classifier output inaccurately high confidence even for input samples that are difficult to classify, which is also known as the over-confidence problem. Even if we collect samples with a confidence of one, their accuracy is not $100\%$ necessarily.

Therefore, various methods have been proposed for calibrating the confidence of NNs~\cite{pmlr-guo17a-oncalib}. Representative methods include temperature scaling, which replaces the softmax function in the final layer of the NN with the softmax with temperature, and histogram binning~\cite{zadrozny2001hist}, which calibrates the confidence by splitting the data into bins. However, these calibration methods require labeled validation data that are different from the training data. Therefore, they are unsuitable for PL, which requires repeating the learning process using a limited number of labeled data.
% ====================================
% Fig Intuitive explanation of the benefit of joint learning
% ====================================
\begin{figure}[t]
    \centering
    \includegraphics[width=1.0\linewidth]{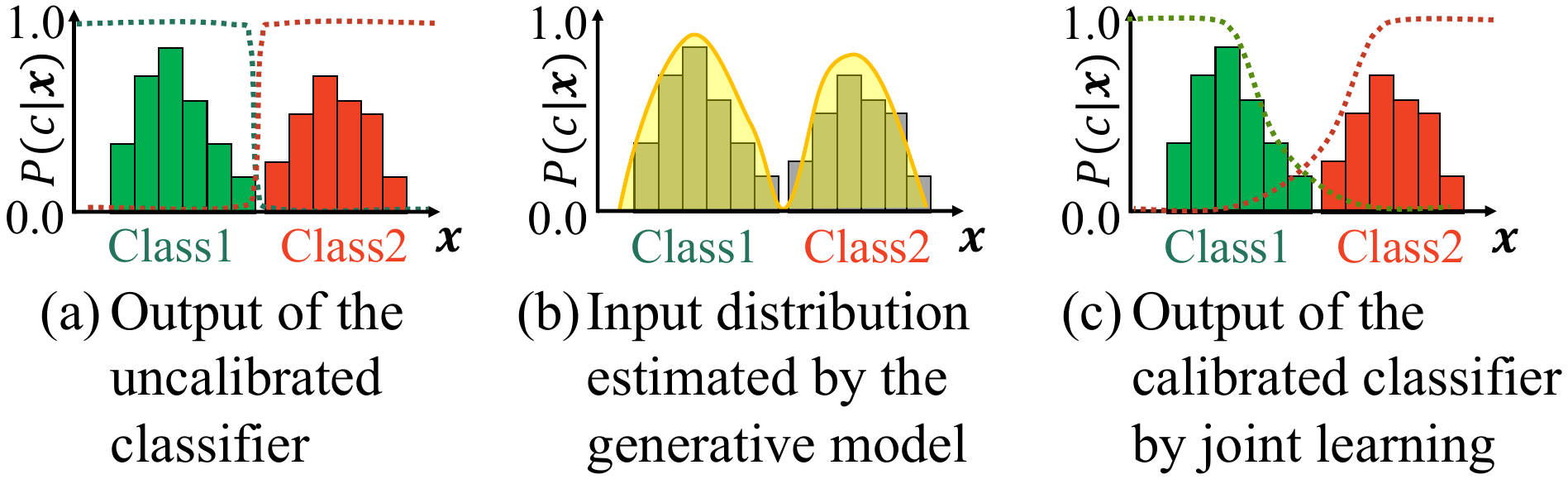}
    \caption{Joint learning of a classifier and generative model and its benefit. (a) Training only the classifier results in wrongly high posterior probability near the class decision boundary. (b) Estimating the input distribution by the generative model allows us to consider the frequency of data occurrences. (c) Joint learning of the classifier and generative model helps to calibrate the confidence.}
    \label{fig:joint_learning_benefit}
\end{figure}
% ====================================

To achieve properly calibrated confidence for PL without using validation data, we propose utilizing the distribution of unlabeled training data via joint learning of a classifier and generative model. Fig.~\ref{fig:joint_learning_benefit} illustrates joint learning of a classifier and generative model and its benefits. Training only the classifier results in wrongly high posterior probability near the decision boundary. By estimating the input data distribution simultaneously using a generative model, the classifier is trained to consider the frequency of data occurrences. This results in lower posterior probabilities for ambiguous data near the decision boundary, thereby leading to calibrated confidence.

In this study, we propose a PL algorithm using an energy-based model (EBM), which is referred to as the energy-based PL (EBPL). In EBPL, an NN-based classifier and an EBM, which is a type of generative model, are jointly trained by sharing their feature extraction parts. This approach enables us to train the classifier by considering the input data distribution, thereby enhancing confidence calibration during NN training in each PL step. By referring to calibrated confidence, we can assign more accurate pseudo-labels, leading to more successful PL.

The main contributions of this paper are as follows.
\begin{itemize}
    \item We propose a PL algorithm that leverages an EBM. Our algorithm involves an NN-based classifier that shares its feature extraction part with an EBM, enabling joint learning of a class decision boundary and input data distribution. This approach helps calibrate confidence during NN training. 
    \item By utilizing EBM, unlabeled data can be effectively used in each step of PL, leading to improved accuracy of the classifier, particularly in the initial learning step.
    \item We demonstrate that the proposed algorithm outperforms the existing confidence-based PL in terms of accuracy and confidence calibration on image datasets with limited labeled data. 
\end{itemize}

\section{Related work}
\label{sec:related}
\subsection{Pseudo-labeling}
\label{ssec:pseudo}
PL is one of the semi-supervised learning methods. In this approach, a classifier is initially trained on labeled data. The output of the classifier itself is then used to assign pseudo-labels to unlabeled data, and the updated data are used for further training. The research carried out by Lee~\cite{lee2013pseudo} is a pioneering effort in the application of pseudo-labeling to deep learning, where the predictive class with the largest posterior probability for an unlabeled sample is considered to be the correct class for that sample. This method can be applied to any network structure and learning method because it can be trained within the supervised learning framework.

Many frameworks have been proposed for PL since it is important to reduce the number of incorrect pseudo-labels and to assign as many correct pseudo-labels as possible in order to improve classification accuracy. For example, uncertainty-aware pseudo-label selection~\cite{DBLP:journals/corr/UPS}, which assigns pseudo-labels by referring to multiple pieces of information output by NNs. Curriculum labeling~\cite{Cascante-Bonilla_Tan_Qi_Ordonez_2021} gradually assigns pseudo-labels from easy to difficult samples. FixMatch~\cite{NEURIPS2020_fixmatch} combines consistency regularization and PL to achieve high classification accuracy while training on a small amount of labeled data. Wang \textit{et al.}~\cite{wang2022debiased} focused on the fact that pseudo-labels are naturally imbalanced even when a model is trained on balanced source data and proposed a debiased learning method with pseudo-labels, based on counterfactual reasoning and adaptive margins. Li \textit{et al.}~\cite{li2023class} adapted a PL approach to semi-supervised federated learning. Liu \textit{et al.}~\cite{liu2023confidence} proposed a confidence-aware PL approach for the weakly supervised visual grounding problem. Li and Lijun~\cite{li2023contactless} applied PL and self-supervised learning to remote photoplethysmography. PL methods using two models, teacher and student models, or combining active learning have also been proposed~\cite{Xie_2020_NoisyS, Pham_2021_MetaPL, Zhang_2022_MIS}. 

\subsection{Energy-based Models}
An EBM is a type of generative model that differs from other models in its approach to probability representation. Rather than directly modeling a normalized probability, an EBM focuses on a negative log-probability, which is also referred to as the energy function. The model's probability density function is derived by normalizing this energy function, expressed as $p_{\bm{\theta}}(\bm{x}) = \exp(-E_{\bm{\theta}}(\bm{x}))/Z_{\bm{\theta}}$. In this formulation, $E_{\bm{\theta}}$ is the energy function parameterized by $\bm{\theta}$, and $Z_{\bm{\theta}}$ is the normalizing constant, computed as $Z_{\bm{\theta}} = \int_{\bm{x}} \exp(-E_{\bm{\theta}}(\bm{x})) \mathrm{d}\bm{x}$. A key challenge in working with EBMs is that this integral for the normalizing term is typically intractable. To estimate it, advanced sampling techniques are often employed, such as the Markov chain Monte Carlo method (MCMC) and stochastic gradient Langevin dynamics (SGLD)~\cite{welling2011bayesian}. 

EBMs are utilized in a vast range of applications such as image generation~\cite{han2019divergence,du2019implicit}, texture generation~\cite{xie2018cooperative}, text generation~\cite{deng2020residual,bakhtin2021residual}, and in other domains including compositional generation, memory modeling, protein design and folding, outlier detection, confidence calibration, adversarial robustness, and semi-supervised learning~\cite{Grathwohl2020JEM}. Their versatility extends to reinforcement learning and continual learning as well. In terms of advancements in EBMs, Du \textit{et al.}~\cite{du2021improved} have notably improved the training stability of these models. They achieved this by introducing a Kullback--Leibler divergence between the MCMC kernel and the model distribution. Furthermore, their research demonstrated that employing data augmentation and multi-scale processing can enhance both the robustness and the quality of generation in EBMs. Salehinejad and Valaee~\cite{salehinejad2021edropout} have applied EBMs innovatively in the context of dropout and pruning within NNs. They used these models to evolve pruning state vectors towards more optimal configurations, which facilitates more efficient pruning of NNs while preserving high levels of classification performance. Additionally, Xu \textit{et al.}~\cite{Xu2022energy} have utilized EBMs to tackle the complex problem of continuous inverse optimal control. This involves learning unknown cost functions across a sequence of continuous control variables, based on expert demonstrations. Their work exemplifies the potential of EBMs in addressing intricate control and optimization challenges.

\subsection{Confidence Calibration}
\label{ssec:confidence}
Confidence calibration is the task of predicting probability estimates that are representative of the true correctness likelihood~\cite{pmlr-guo17a-oncalib}. In classification using an NN, the objective is to ensure that the posterior probability of the predicted class, i.e., the confidence, aligns with the actual classification rate. When performing PL, it is important to have a calibrated classifier that can effectively utilize confidence information. Specifically, the NN should output a lower confidence for input samples that are difficult to classify and output a higher confidence for samples that clearly belong to a particular class.

Confidence calibration methods can be classified into two approaches: train-time calibration, which is performed during training, and post-hoc calibration, which is performed after training has been completed~\cite{patra2023calibrating,munir2023bridging}. The main difference between them is whether the validation data is used to calibrate the confidence.\par

Post-hoc calibration has been a primal approach, and many methods have been proposed~\cite{pmlr-guo17a-oncalib,zadrozny2001hist,naeini2015obtaining,platt1999platt,zadrozny2002transforming}. Among such methods, temperature scaling~\cite{pmlr-guo17a-oncalib} is considered simple and effective. It uses a softmax function with temperature instead of the usual softmax in the final layer of the NN and tunes the temperature parameter to optimize the negative likelihood. The computational cost is relatively low, but since the temperature parameter is tuned after training using validation data different from training data, it may not work correctly if the validation data is biased or its number is small. Seo \textit{et al.}~\cite{seo2019learning} developed a specialized confidence calibration technique for medical image segmentation employing deep NNs, enhancing the precision and reliability of segmentation tasks in medical applications. Further extending the scope of confidence calibration, Wang \textit{et al.}~\cite{wang2021confident} proposed a method specifically tailored for graph neural networks (GNNs). This innovation addresses the unique challenges in calibrating confidence in GNNs, which are increasingly used for complex data structures like social networks, molecular structures, and recommendation systems. Fernando and Tsoko~\cite{fernando2021dynamically} explored the challenges of imbalanced data classification. They demonstrated the effectiveness of a class rebalancing strategy using a class-balanced dynamically weighted loss function. This approach significantly improves the calibration of a classifier's confidence in scenarios where data is unevenly distributed across different classes, a common issue in real-world datasets.

One of the methods of train-time calibration is the simultaneous learning of a classifier with a generative model, such as the joint energy-based model (JEM)~\cite{Grathwohl2020JEM}. JEM combines an NN-based classifier and an EBM that shares their feature extraction parts. By estimating the input data distribution at the same time as the classifier's training, the classifier considers the frequency of data occurrence and obtains calibrated confidence. By performing train-time confidence calibration, the need for a validation dataset is eliminated. Mixup~\cite{zhang2018mixup} is another technique that has been reported to be effective for confidence calibration~\cite{thulasidasan2019mixup-calib}, which convexly combines two images and their labels in the training data at a specific rate; however, its effectiveness may be limited when the number of labeled data is insufficient.

The algorithm of EBPL employs a train-time calibration method called the hybrid model~\cite{hayashi2023hybrid}, which jointly trains a classifier and EBM. The difference from JEM is that the classifier and EBM parameters are independent, and their correlation can be arbitrarily determined by the final term of the loss function, which leads to better classification accuracy and calibrated confidence. 

\section{Energy-based Pseudo Labeling (EBPL)}
\label{sec:proposed}
The proposed EBPL achieves accurate PL by utilizing calibrated confidence when assigning pseudo-labels. To calibrate the confidence, EBPL employs the hybrid model~\cite{hayashi2023hybrid} as the classifier, which calibrates the confidence output by an NN by simultaneously performing distribution estimation with EBM during the classifier's training.

\subsection{Algorithm of EBPL}
\label{ssec:pseudo-algo}
% ====================================
% Fig Algorithm of EBPL
% ====================================
\begin{figure}[t]
    \centering
    \includegraphics[width=1.0\linewidth]{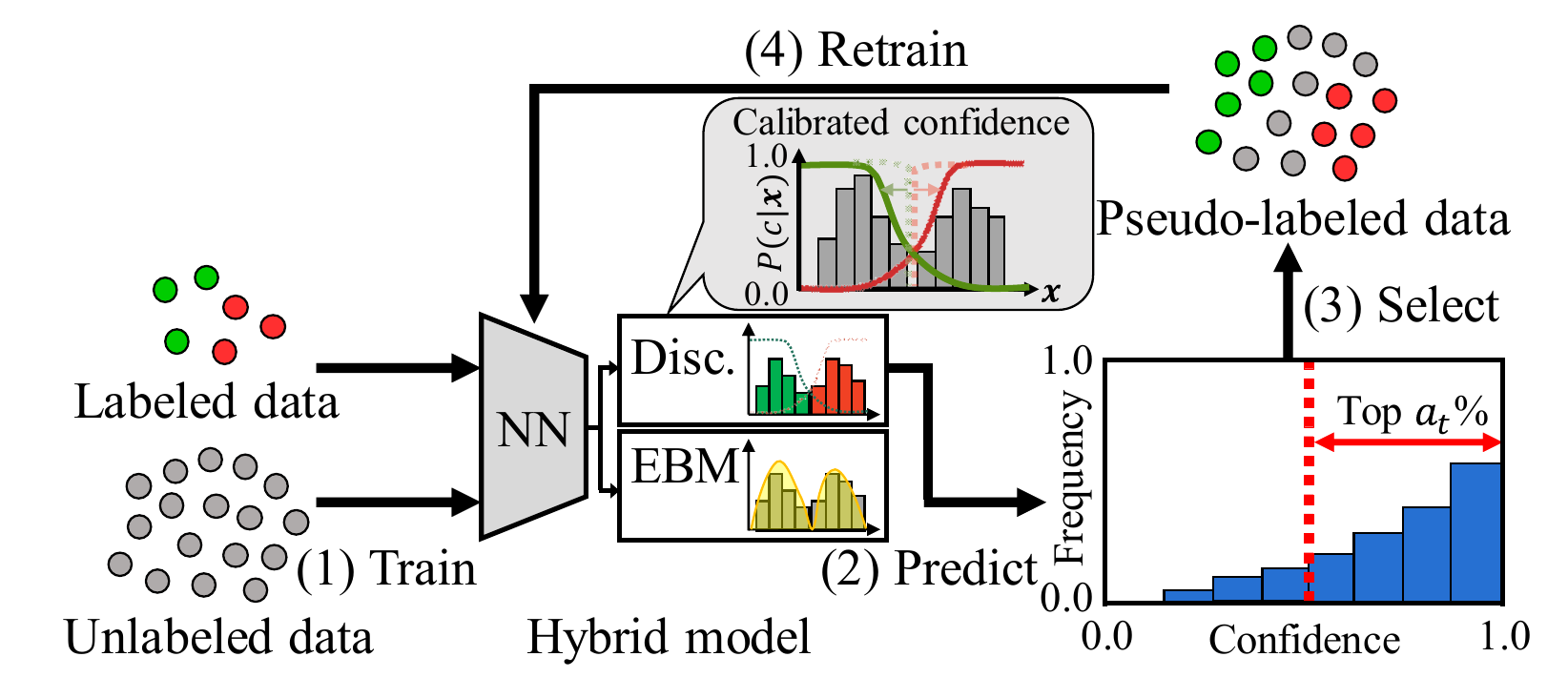}
    \caption{Overview of the EBPL algorithm.}
    \label{fig:EBPL_algorithm}
\end{figure}
% ====================================
Fig.~\ref{fig:EBPL_algorithm} outlines the algorithm of EBPL. First, the hybrid model is trained on both labeled and unlabeled data once, and then pseudo-labels are assigned to unlabeled data step by step until pseudo-labels are assigned to all the unlabeled data. There are multiple pseudo-label assignment steps. In each pseudo-label assignment step $t~(t = 1, \ldots, T)$, the confidence scores for the unlabeled data are sorted in descending order, and pseudo-labels are assigned to data with the top $\alpha_t$\% of confidence scores, where $\alpha_t$ increases according to step $t$ so that pseudo-labels are assigned to all unlabeled data in the final step $T$ (i.e., $\alpha_T=100$\%). After each pseudo-label assignment step, the hybrid model is trained on labeled data, pseudo-labeled data, and unlabeled data. The pseudo-labels assigned in the previous step are discarded, and new pseudo-labels are assigned. This prevents the accumulation of false pseudo-labels in the early stages of learning. 

The algorithm of EBPL is inspired by curriculum labeling~\cite{Cascante-Bonilla_Tan_Qi_Ordonez_2021}, and there are three main differences between EBPL and curriculum labeling. First, curriculum labeling uses only labeled data for the initial training, while EBPL uses all training data including unlabeled data for the initial training because unlabeled data can also be used in the hybrid model to estimate the input data distribution. Second, while the original curriculum labeling initializes all model parameters at the beginning of each step, EBPL does not initialize to minimize the learning cost. Third, in EBPL, the posterior probabilities are employed for pseudo-labels; namely, the soft labels are used instead of hard labels to preserve calibrated confidence information for continuity into the next training step.

\subsection{Hybrid Learning of a Classifier and an Energy-based Model}
\label{ssec:hybrid}
\begin{figure}[t]
    \centering
    \includegraphics[width=1.01\linewidth]{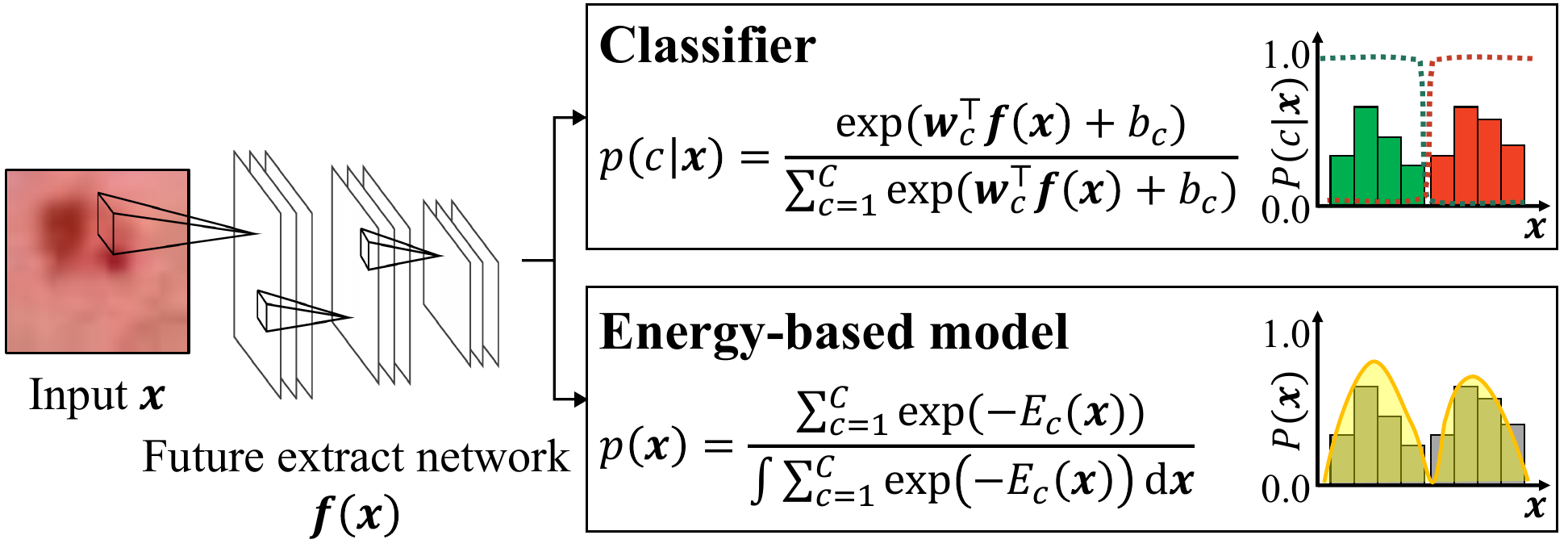}
    \caption{Structure of the hybrid model employed in EBPL.}
    \label{fig:hybrid}
\end{figure}

Fig.~\ref{fig:hybrid} shows the network structure of the hybrid model. The hybrid model consists of a feature extraction network $\bm{f}(\bm{x})$ for the input sample $\bm{x}$, a classifier that outputs the posterior $p(c \mid \bm{x})$ of class $c \in \{1, \ldots, C\}$, and an EBM that estimates the probability density function $p(\bm{x})$. 
% ====================================
% Fig: Structure of the hybrid model
% ====================================
% ====================================
The classifier outputs the posterior probability $p(c \mid \bm{x})$ of class $c$ based on a fully-connected layer with a weight $\bm{w}_c$ and bias $b_c$ and a softmax function. 
\begin{equation}
\label{eq:Softmax}
p(c \mid \bm{x}) = \frac{\exp({\bm{w}_{c}}^\top\bm{x} + b_c)}{\sum\limits^{C}_{c'=1}\exp({\bm{w}_{c'}}^\top\bm{x} + b_{c'})}
\end{equation}
In the EBM, we assume that the feature vector output by $\bm{f}(\bm{x})$ follows the Gaussian distribution for each class, and the input distribution $p(\bm{x})$ is modeled as follows: 
\begin{equation}
\label{eq:EBM}
{p(\bm{x})}=\frac
{\sum\limits_{c=1}^C\exp\left(-E_c(\bm{x})\right)}
{\int \sum\limits_{c=1}^C\exp\left(-E_c(\bm{x})\right)
\mathrm{d}\bm{x}},
\end{equation}
where $E_c(\bm{x}) = \frac{1}{2}(\bm{f}(\bm{x})-\bm{\mu}_{c})^\top\bm{\Sigma}^{-1}(\bm{f}(\bm{x})-\bm{\mu}_{c})$ and $\bm{\mu}_{c}$ and $\bm{\Sigma}$ represent the mean vector for each class and the covariance matrix shared among all the classes, respectively. By jointly learning ${\bm{\mu}}_{c}$ and $\bm{\Sigma}$ with the parameters of the feature extraction network $\bm{f}(\bm{x})$, the input data distribution is estimated. 

By training the classifier and EBM with the same feature extraction network, the model obtains a feature space that reflects the high discriminative ability of the classifier and the high confidence calibration ability of the EBM. When training only a classifier, it may achieve high classification accuracy, but as mentioned earlier, it may suffer from the over-confidence problem for ambiguous data near the discrimination boundary. By incorporating the EBM into the training, the NN not only performs classification but also estimates the input data distribution simultaneously. This enables the NN to learn to output low confidence for infrequent data that may be near the boundaries. As a result, the feature space benefits from the complementary characteristics of the classifier and EBM.

In the training of the hybrid model, we minimize the following loss function for a partially labeled dataset $\{\bm{x}_n, c_n\}_{n=1}^{M}\cup\{\bm{x}_n\}_{n=M+1}^{N}$, where $N$ is the total number of training data, and $M$~($M\leq N$) is the number of labeled data.
\begin{flalign}
\label{eq:loss_function}
\mathcal{L} &\!=\! 
    -\sum_{n=1}^M\log p(c_n|\bm{x}_n) 
    -\sum_{n=1}^{N}\log p(\bm{x}_{n}) \nonumber \\
    &\;\;+\! \lambda\! \sum_{c=1}^C \left\{\!\left\|\bm{w}_{c} \!\!-\!\! \bm{{\bf \Sigma}^{-1}\!\mu}_{c}\right\|^2 \!\!+\!\! \left\|b_c \!+\! \frac{1}{2}\bm{\mu}_{c}^\top\!{\bf \Sigma}^{-1}\!\bm{\mu}_{c}\right\|^2\!\right\}
\end{flalign}
where the first term on the right-hand side corresponds to the cross-entropy for the classifier, and the second term is the negative log-likelihood for the EBM. The third term is responsible for correlating the parameters of the classifier and the EBM, where $\lambda$ is a hyperparameter. The gradient of the second term cannot be computed in a straightforward manner because it involves an intractable normalizing term; therefore, it is computed using stochastic gradient Langevin dynamics (SGLD) sampling~\cite{welling2011bayesian}. The derivative of $\log p(\bm{x}_n)$ with respect to the EBM parameter $\bm{\theta}$ can be calculated as follows:
\begin{equation}
    \frac{\partial \log p(\bm{x}_n)}{\partial \bm{\theta}} \!=\! \mathbb{E}_{p(\bm{x'})}\left[\frac{\partial E_{\mathrm{total}}(\bm{x'})}{\partial \bm{\theta}} \right] \!-\! \frac{\partial E_{\mathrm{total}}(\bm{x}_n)}{\partial \bm{\theta}},
\end{equation}
where $E_{\mathrm{total}}(\bm{x}_n) = -\log\sum\limits_{c=1}^C\exp(-E_c(\bm{x}_n))$. The expectation over $p(\bm{x}_n)$ is calculated via sampling from the estimated distribution using SGLD.
\begin{align}
    \bm{x}_{0} &\sim p_{0}(\bm{x}), \nonumber \\
    \bm{x}_{i+1} &= \bm{x}_{i} - \frac{\alpha}{2} \frac{\partial E_\mathrm{total}(\bm{x}_i)}{\partial \bm{x}_i} + \epsilon, \quad \epsilon \sim \mathcal{N}(0, \alpha),
\end{align}
where $p_{0}$ is the initial distribution that is typically a uniform distribution and $\alpha$ is the step size. The third term of Eq.~(\ref{eq:loss_function}) controls the strength of the connection between the classifier and EBM. The details are described in the Appendix.

\section{Experiment}
\label{sec:exp}
To assess the effectiveness of EBPL, we conducted image classification experiments. In the experiments, we evaluated the classification accuracy and confidence calibration ability of EBPL in semi-supervised classification tasks. We also conducted both quantitative and qualitative analyses of the results to provide a comprehensive evaluation of the proposed method.

\subsection{Experimental Conditions}
\label{ssec:subhead}
We used seven publicly accessible image datasets. CIFAR-10~\cite{krizhevsky2009cifar10} consists of $32 \times 32$ generic object images in ten classes with 50,000 training images and 10,000 test images. SVHN~\cite{netzer2011reading} contains $32 \times 32$ color images of house numbers collected from Google Street View images. There are 73,257 digits for training and 26,032 digits for testing. Fashion-MNIST~\cite{xiao2017} comprises $28 \times 28$ grayscale images of 70,000 fashion products from 10 categories. The dataset is split into a training set of 60,000 images and a test set of 10,000 images. We also employed four medical datasets from MedMNIST~\cite{medmnist}, including Blood-MNIST, OrganA-MNIST, OrganC-MNIST, and OrganS-MNIST. Blood-MNIST is a $28 \times 28$ color image dataset in eight classes with 17,092 images. OrganA-MNIST, OrganC-MNIST, and OrganS-MNIST are the $28 \times 28$ grayscale image datasets in 11 classes and consist of 58,830, 23,583, and 25,211 images, respectively. To utilize these datasets for a semi-supervised classification task, the number of labeled data was randomly reduced to about 1\% of the original one, reserving the remainder as unlabeled data. This random subsampling was performed after isolating 30\% of the original training data for the validation set. Additionally, to investigate the influence of labeled image quantity, we conducted experiments on Blood-MNIST by varying the number of labeled images for each class from one to five.

We used the accuracy, F-score, and expected calibration error~(ECE)~\cite{naeini2015obtaining} as evaluation criteria. The accuracy is the ratio of the number of correct predictions to the total number of test samples, and the F-score is defined by the harmonic mean of precision and recall. ECE is quantified as $\mathrm{ECE} = \sum\limits^M_{m=1}\frac{|B_m|}{n}\left|\mathrm{acc}(B_m)-\mathrm{conf}(B_m)\right|$, where $n$ denotes the total number of samples, and $B_m$ represents the subset of samples whose predicted confidence scores lie within the interval $I_m = (\frac{m-1}{M}, \frac{m}{M}]$. The accuracy function $\mathrm{acc}(B_m)$ for each bin $B_m$ is defined as $\mathrm{acc}(B_m) = \frac{1}{|B_m|}\sum\limits_{i \in B_m}1(\hat{\bm{c}}_i=\bm{c}_i)$, where $\hat{\bm{c}}_i$ and $\bm{c}_i$ are the predicted and actual class labels, respectively, for the $i$-th sample in $B_m$. The average confidence $\mathrm{conf}(B_m)$ is calculated as $\mathrm{conf}(B_m) = \frac{1}{|B_m|}\sum\limits_{i \in B_m}\hat{r}_i$, with $\hat{r}_i$ signifying the predicted confidence for each sample in $B_m$. ECE measures the degree of agreement between confidence and the probability of correctness, and it is calculated by the expected difference between the actual accuracy and the confidence calculated for each bin of confidence. Since the evaluation index varied depending on the random seed of the experimental program to select the labeled data, we calculated the mean value and standard deviation for five seeds.

We used the wide residual network~\cite{BMVC2016_87_wideresnet} for the feature extraction network. We set the number of epochs for the initial training to $120$, the number of PL steps to $T=4$, and the number of training epochs in each step to $20$. The result for the test data was calculated using the weights with the highest validation accuracy throughout all training steps. We used Adam optimizer with a learning rate of $0.0001$. The hyperparameter $\lambda$ was set to $0.001$. The PL rate was set based on $\alpha_t = 100t/T$ ($t = 1, \ldots, T$).

For the comparative study, we set a baseline method based on the curriculum labeling~\cite{Cascante-Bonilla_Tan_Qi_Ordonez_2021}. For this baseline, we used the same feature extraction network, hyperparameters such as the number of training epochs, number of PL steps, and learning rate as EBPL. The main difference from EBPL is two-fold: one with the last layer of the classifier and the other with the weight initialization. The baseline classifier has only a fully-connected layer with the softmax function in the last layer and does not have any input distribution estimator. Furthermore, following the original curriculum labeling algorithm, the baseline method randomly initializes the NN weights at the beginning of each PL step to reduce concept drift and confirmation bias~\cite{DBLP:journals/corr/confirmation}. 

\subsection{Quantitative Evaluation}
\label{ssec:quantitative}
% ====================================
% Table Accuracy and ECE
% ====================================
\begin{table*}[t]
\caption{Accuracy, F-score, and expected calibration error~(ECE). Note that only about 1\% of labeled data for each class were randomly sampled from the original training dataset. The scores are represented in percentages. 
% The values in parentheses represent the standard deviations.
}
\scalebox{0.93}{
% \begin{tabular}{llllllllllllllllll}
\begin{tabular}{@{}ccccccccc@{}}
\toprule
\multicolumn{1}{c}{} &         & {\begin{tabular}{c}CIFAR-10~\cite{krizhevsky2009cifar10}\\350 labels\end{tabular}} & {\begin{tabular}{c}SVHN~\cite{netzer2011reading}\\510 labels\end{tabular}} & {\begin{tabular}{c}Fashion~\cite{xiao2017}\\420 labels\end{tabular}} & {\begin{tabular}{c}Blood~\cite{medmnist}\\80 labels\end{tabular}} & {\begin{tabular}{c}OrganA~\cite{medmnist}\\352 labels\end{tabular}} & {\begin{tabular}{c}OrganC~\cite{medmnist}\\132 labels\end{tabular}} & {\begin{tabular}{c}OrganS~\cite{medmnist}\\143 labels\end{tabular}} \\  \cmidrule(l){3-9}
Method & Pseudo-label & \multicolumn{7}{c}{Accuracy (\%)$\uparrow$} \\ \cmidrule(r){1-2} \cmidrule(l){3-9}
Baseline & w/o & 36.17 (1.354)  & 53.34 (13.45) & 78.35 (1.440)  & 74.07 (2.829)  & 80.20 (1.334)  & 63.08 (1.957)  & 49.81 (2.894) \\
 & Hard & 39.98 (1.920)  & 69.29 (13.23)  & 81.75 (0.8697)  & 78.88 (1.102)  & 83.75 (0.526)  & 72.79 (2.342)  & 56.86 (4.223) \\
 & Soft & 41.42 (1.722)  & 66.90 (12.86)  & 81.72 (0.7066)  & 78.53 (1.407)  & 84.02 (1.218)  & 71.59 (3.078)  & 58.20 (3.162) \\ %\hdashline
EBPL & w/o & 50.29 (0.7313)  & 67.26 (3.262)  & 82.97 (0.6444)  & 79.26 (2.929)  & 85.95 (0.6750)  & 75.73 (1.831)  & 61.92 (0.9848) \\
 & Hard & \textbf{53.89} (1.333)  & \textbf{79.39} (4.074)  & 82.92 (0.4625)  & 80.51 (2.655)  & 86.76 (0.8712)  & 78.97 (1.797)  & \textbf{64.10} (1.810) \\
 & Soft & 53.63 (1.371)  & 77.58 (3.935)  & \textbf{83.12} (0.6995)  & \textbf{80.93} (2.218)  & \textbf{86.99} (0.8568)  & \textbf{79.55} (1.186)  & 63.30 (2.046) \\  \cmidrule(l){3-9}
 &  & \multicolumn{7}{c}{F-score (\%)$\uparrow$}\\  \cmidrule(l){3-9}
Baseline & w/o & 35.90 (1.285)  & 51.96 (13.24) & 78.34 (1.393)  & 70.53 (2.837)  & 80.40 (1.316)  & 63.34 (1.598)  & 47.37 (2.270) \\
 & Hard & 38.79 (1.933)  & 67.22 (13.40)  & 81.51 (0.8316)  & 75.52 (1.483)  & 83.56 (0.6996)  & 71.26 (2.518)  & 51.71 (4.638) \\
 & Soft & 40.25 (1.741)  & 64.98 (13.34)  & 81.51 (0.8657)  & 75.05 (1.543)  & 83.80 (1.372)  & 70.34 (2.726)  & 53.23 (3.477) \\ %\hdashline
EBPL & w/o & 49.83 (1.078)  & 65.52 (3.439)  & 82.69 (0.7169)  & 75.85 (3.305)  & 85.63 (0.6653)  & 74.14 (1.860)  & 56.18 (0.9768) \\
 & Hard & 52.66 (1.755)  & \textbf{77.69} (4.528)  & 82.47 (0.6404)  & 77.11 (2.916)  & 86.64 (0.9494)  & 76.95 (1.492)  & \textbf{56.96} (1.819) \\
 & Soft & \textbf{53.04} (1.637)  & 75.87 (4.021)  & \textbf{82.73} (1.017)  & \textbf{77.88} (2.259)  & \textbf{86.88} (0.8743)  & \textbf{77.04} (1.174)  & 56.69 (2.352) \\  \cmidrule(l){3-9}
 &  & \multicolumn{7}{c}{ECE (\%)$\downarrow$} \\  \cmidrule(l){3-9}
Baseline & w/o & 55.04 (2.068)  & 40.97 (12.05)  & 19.16 (0.9510)  & 20.23 (2.731)  & 16.27 (1.788)  & 30.62 (1.431)  & 42.50 (3.380) \\
 & Hard & 42.09 (1.824)  & 17.62 (9.871)  & 12.33 (0.9162)  & 15.58 (1.230)  & 7.502 (0.7183)  & 14.55 (2.347)  & 25.87 (3.681) \\
 & Soft & 35.53 (1.566)  & 15.04 (8.053)  & 11.95 (0.7853)  & 14.58 (2.641)  & 6.274 (1.388)  & 11.80 (1.452)  & 20.14 (1.516) \\ %\hdashline
EBPL & w/o & \textbf{15.18} (3.891)  & 5.167 (2.822)  & \textbf{11.35} (0.8308)  & \textbf{8.996} (3.031)  & \textbf{5.500} (0.6261)  & \textbf{5.771} (1.811)  & \textbf{10.03} (1.511) \\
 & Hard & 30.20 (2.889)  & 13.39 (3.881)  & 13.04 (0.8979)  & 12.57 (4.701)  & 6.955 (0.4265)  & 11.96 (3.282)  & 21.91 (2.993) \\
 & Soft & 19.93 (3.661)  & \textbf{4.941} (3.566)  & 12.48 (1.055)  & 9.846 (3.043)  & 5.604 (1.204)  & 9.587 (1.085)  & 18.87 (2.843) \\
 \bottomrule
\end{tabular}}
\label{table:testdata}
\end{table*}
% ====================================
Table~\ref{table:testdata} shows the accuracy, F-score, and ECE for each dataset. Broadly speaking, EBPL outperformed the baseline method in terms of all the evaluation criteria. This is because EBPL calibrated the confidence based on the joint learning of the classifier and EBM, thereby improving the accuracy of PL. The details are investigated in subsequent analyses. Furthermore, soft PL demonstrated better ECE than hard PL on all datasets, reinforcing the validity of the use of soft pseudo-labels in EBPL. 

% ====================================
% Fig Accuracy and ECE for each number of labels per class
% ====================================
\begin{figure}[t]
    \centering
    \includegraphics[width=1.0\linewidth]{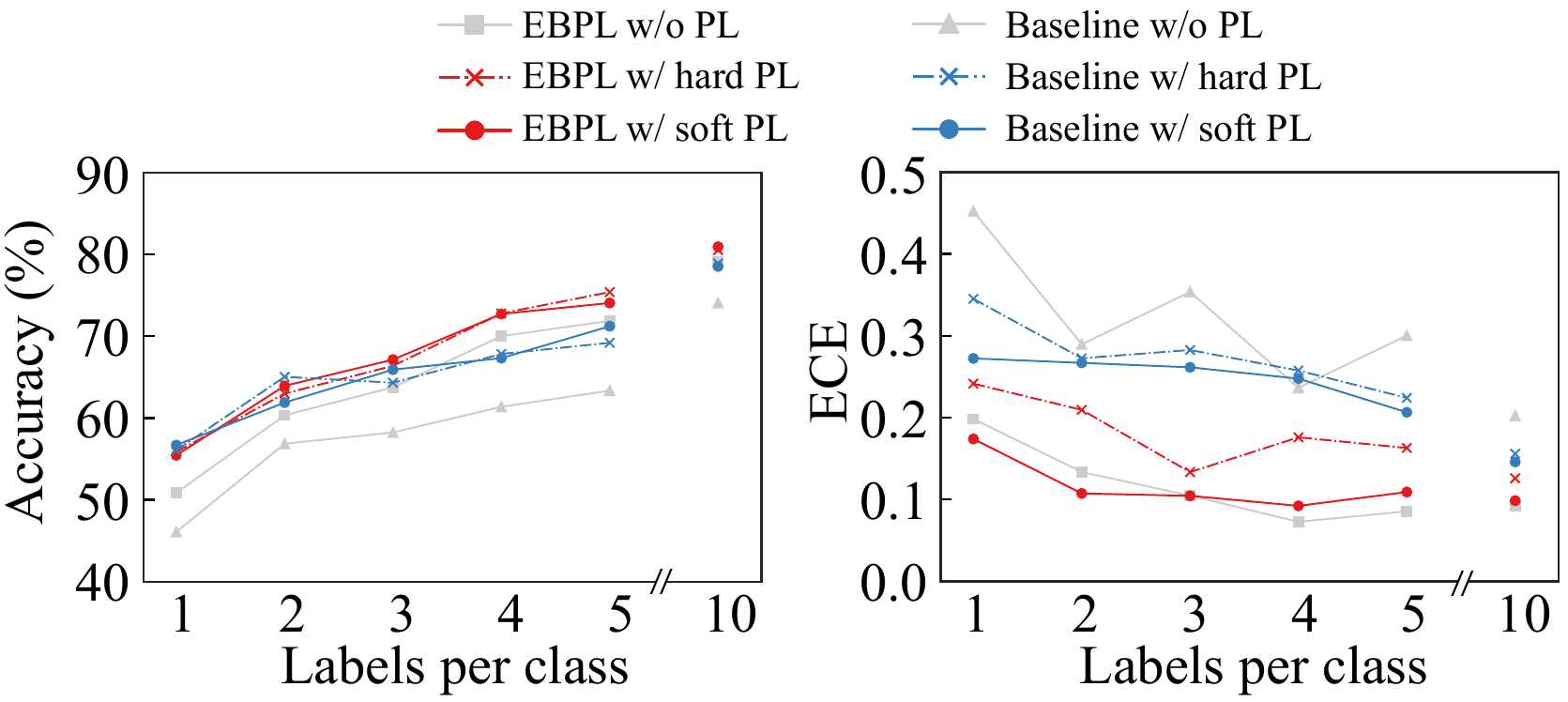}
    \caption{Accuracy and ECE for each method when varying the number of labeled data for each class.}
\label{fig:acc_ece_labels_per_class}
\end{figure}
% ====================================
Fig.~\ref{fig:acc_ece_labels_per_class} shows the changes in accuracy and ECE based on the variation in the number of labeled data for each class on the Blood-MNIST dataset. Even with an extremely small number of labeled data, EBPL outperformed the baseline method. In particular, the differences in ECE are remarkable, demonstrating the strong confidence calibration capability of EBPL. 

% ====================================
% Fig Reliability diagram
% ====================================
\begin{figure}[t]
    \centering
    \includegraphics[width=1.0\linewidth]{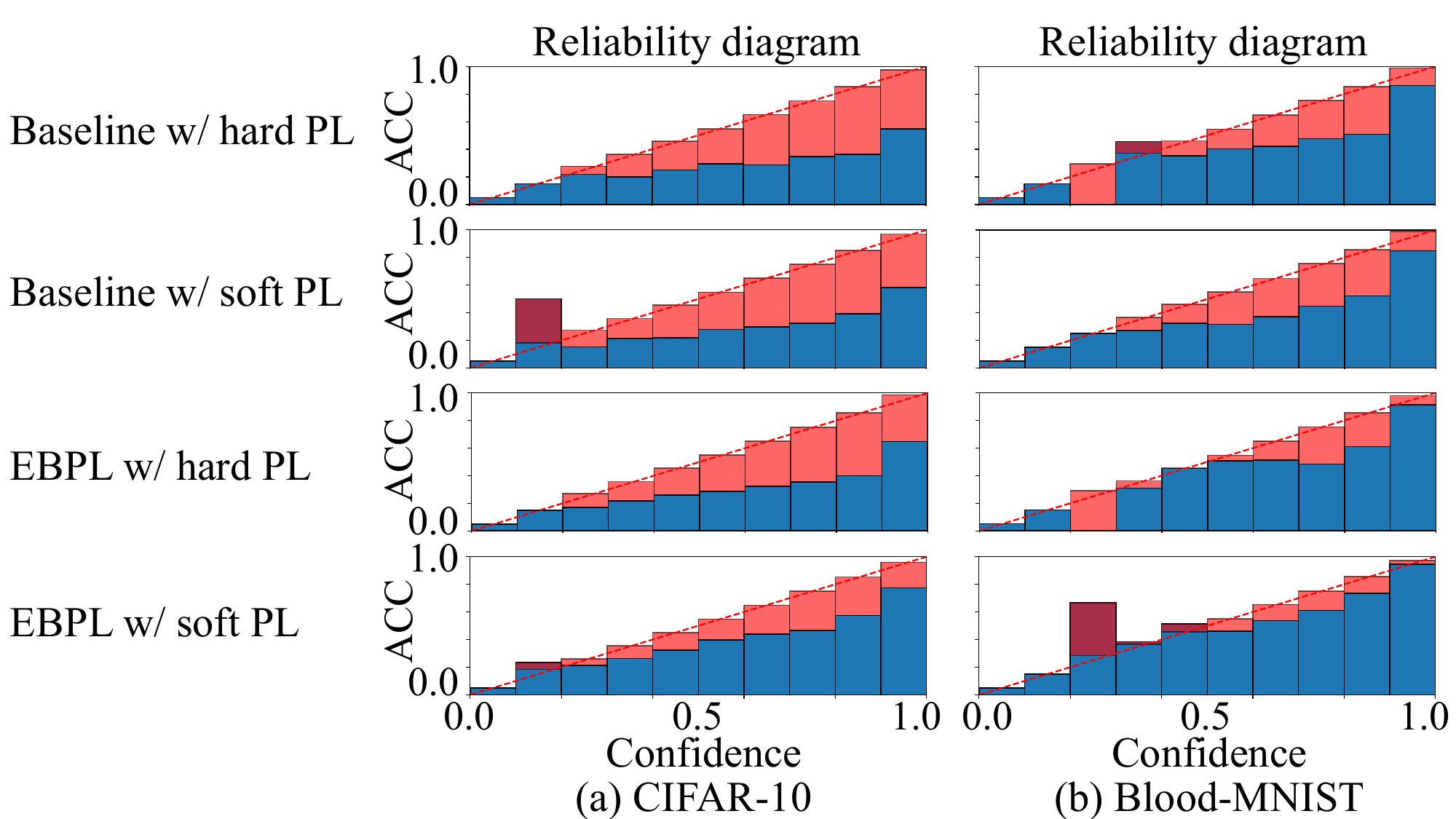}
    \caption{Reliability diagram. Blue and red bars show the actual average accuracy in each bin and the gap to the ideal, respectively. The diagonal red dashed lines show the perfect calibration.}
    \label{fig:reliability_diagram}
\end{figure}
% ====================================
Fig.~\ref{fig:reliability_diagram} shows the reliability diagram, which visualizes how the confidence calibration is achieved for each range of predicted confidence. In this diagram, the average accuracy is calculated for samples whose predicted confidence falls into each interval obtained by dividing [0, 1] into fixed-length segments, and the gap between the actual accuracy and the ideal is displayed. EBPL showed better-calibrated confidence compared to the baseline, and the use of soft PL tended to result in even better performance.

% ====================================
% Fig Pseudo-labeling accuracy at each step
% ====================================
\begin{figure}[t]
    \centering
    \includegraphics[width=1.0\linewidth]{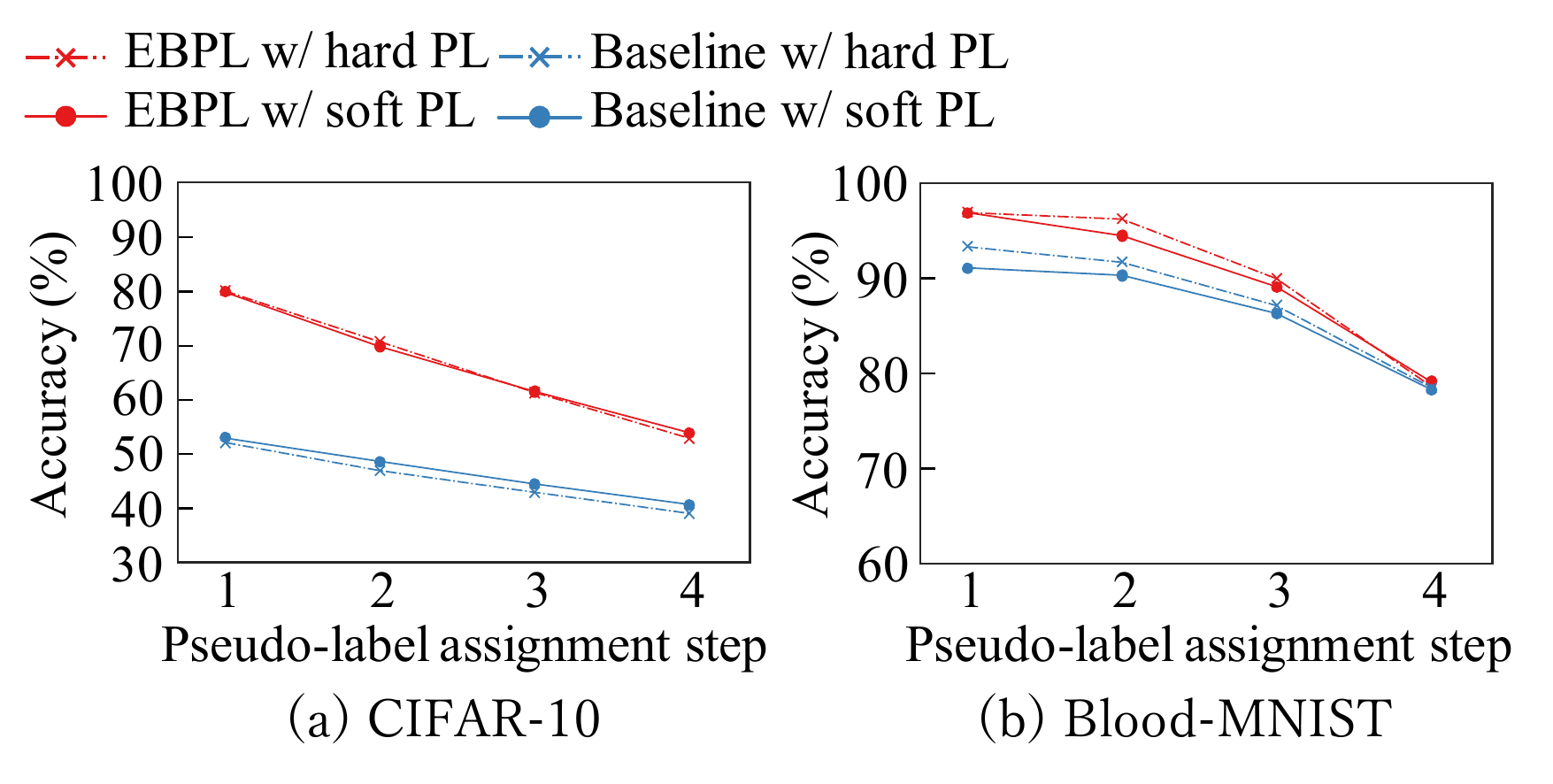}
    \caption{Accuracy of pseudo-labeling at each pseudo-label assignment step.}
    \label{fig:acc_for_each_step}
\end{figure}
% ====================================
The accuracy of PL at each pseudo-label assignment step is shown in Fig.~\ref{fig:acc_for_each_step}, where the accuracy is calculated by comparing assigned pseudo-labels to the ground truth labels. For soft PL, we used the maximum value of the pseudo-label. Compared to the baseline, EBPL demonstrated higher PL accuracy throughout the entire training process. In both methods, the PL accuracy was the highest at step 1 and gradually decreased as the step progressed, likely due to the assignment of pseudo-labels to data with low confidence. The higher PL accuracy of EBPL is attributed to the improvement in discrimination accuracy and confidence calibration using the hybrid model.

\subsection{Qualitative Evaluation}
\label{ssec:qualitative}
% ====================================
% Fig Examples of misclassified images
% ====================================
\begin{figure}[t]
    \centering
    \includegraphics[width=1.0\linewidth]{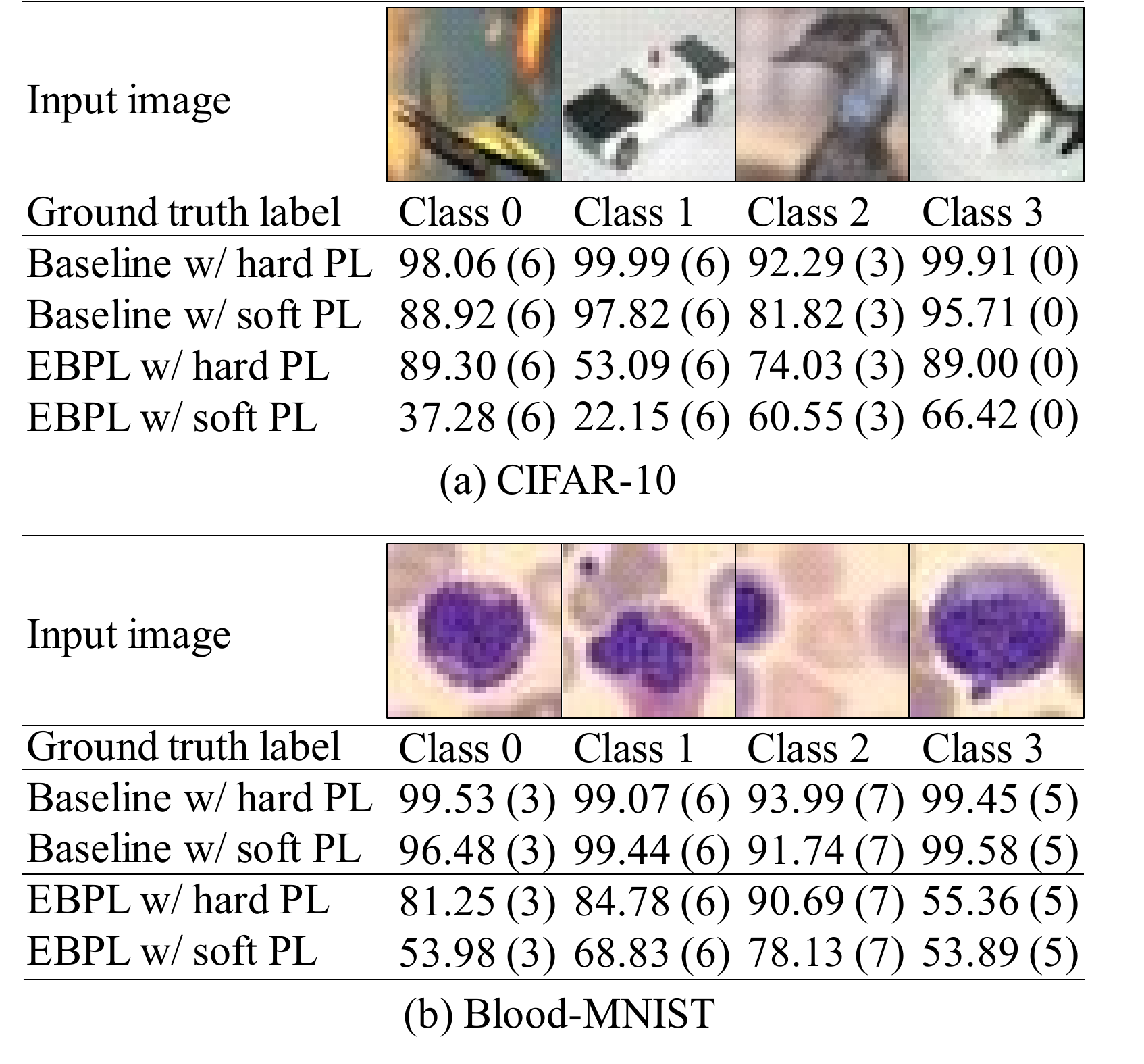}
    \caption{Examples of misclassified images and their predicted confidences. The numbers below the images represent the predicted confidence of each method in percentages. The parenthesized numbers are the predicted classes.}
    \label{fig:misclassified_examples}
\end{figure}
% ====================================
Examples of misclassified images and their corresponding predicted confidences and classes are shown in Fig.~\ref{fig:misclassified_examples}. The confidence score is a measure of the certainty of the classification result, and therefore it should be low for misclassified images. In both datasets, the baseline method produced higher confidence scores for the incorrectly classified samples, while EBPL produced lower confidence scores for those samples. This suggests that EBPL was successful in calibrating the model's confidence, reducing the confidence score for samples with low prediction certainty.

% ====================================
% Fig t-SNE visualization
% ====================================
\begin{figure}[t]
    \centering
    \includegraphics[width=1.0\linewidth]{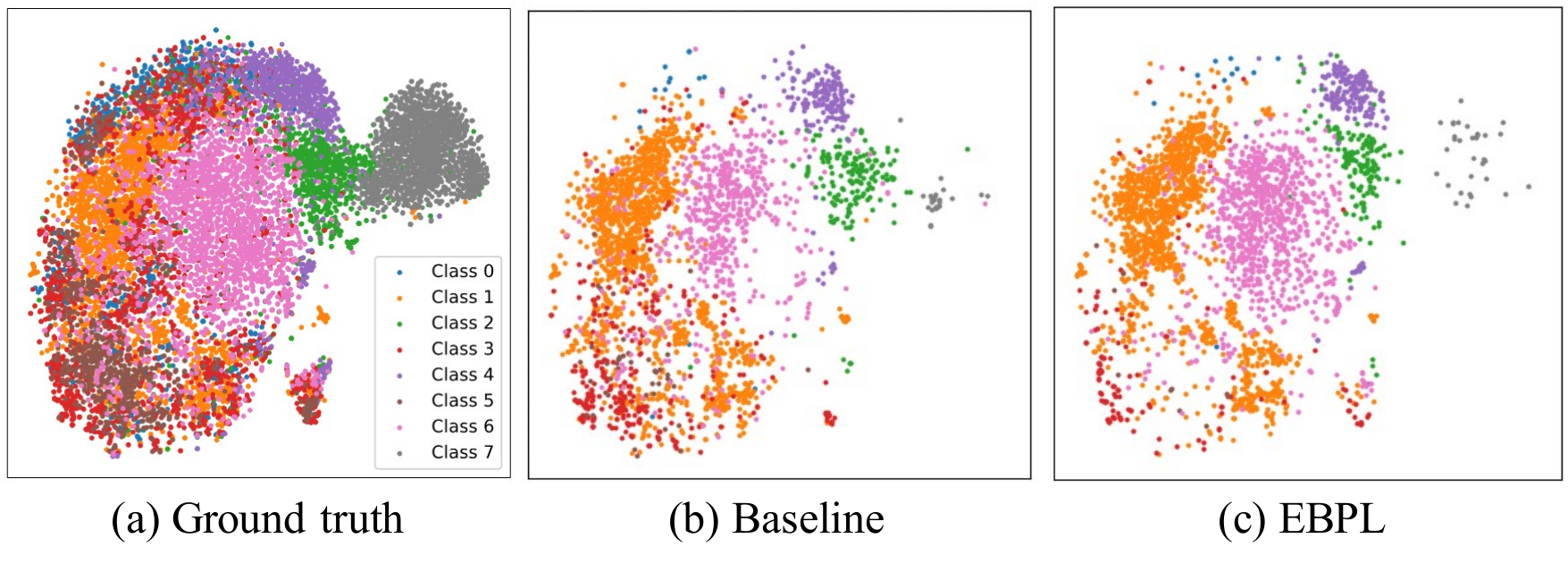}
    \caption{Distributions of the training samples in the Blood-MNIST dataset visualized using $t$-SNE. (a) All training samples, colored according to their ground-truth labels. (b) and (c) Samples selected by the baseline and EBPL methods, respectively, during the first pseudo-label assignment step, with colors representing the assigned pseudo-labels.}
    \label{fig:tSNE_visualization}
\end{figure}
% ====================================
Distributions of the training samples in the Blood-MNIST dataset are shown in Fig.~\ref{fig:tSNE_visualization}. In each figure, the training samples are visualized by compressing the input images into two dimensions by $t$-SNE. Fig.~\ref{fig:tSNE_visualization}(a) shows all the training samples colored according to the ground-truth labels. Figs.~\ref{fig:tSNE_visualization}(b) and (c) are the samples selected by the baseline and EBPL methods, respectively, during the first pseudo-label assignment step, with colors representing the assigned pseudo-labels. Comparing the ground-truth labels and pseudo-label assignments, it can be observed that both the baseline and EBPL methods assign almost correct pseudo-labels. However, EBPL selected less biased samples than the baseline, such that the selected samples are uniformly distributed in each class, particularly in Classes 6 and 7. This is presumably because EBPL acquired the distribution of the input data by simultaneously learning the classifier and the energy-based model, and was able to output appropriate confidence values when assigning pseudo-labels. The results suggested the effectiveness of simultaneous learning of a classifier and energy-based model for pseudo-label selection.

\section{Conclusion}
\label{sec:conclusion}
In this study, we proposed a PL method based on an EBM called EBPL and applied it to semi-supervised image classification. In EBPL, unlabeled data are effectively used for training by assigning pseudo-labels to them in the order of their corresponding confidence scores estimated by the classifier. To obtain better-calibrated confidence scores, an NN-based classifier and an EBM are jointly trained by sharing their feature extraction parts. This approach enables the model to learn both the class decision boundary and input data distribution and helps to calibrate the confidence during network training. The experimental results demonstrated that EBPL showed higher accuracy and better-calibrated confidence compared to the existing confidence-based PL. We also confirmed that the proposed EBPL is more accurate in assigning pseudo-labels in all pseudo-label assigning steps compared to the existing curriculum labeling algorithm.

The limitations of EBPL include the computational cost and the size of input images. Since EBPL employs an energy-based model that requires sampling-based gradient estimation during training, it incurs more training costs than a normal classifier. Moreover, as learning an EBM becomes more difficult with large input image sizes, the experiments in this paper are limited to ones using relatively small image sizes, such as the CIFAR-10 with an image size of $32 \times 32$.

In future work, we will investigate further potential uses of the EBM jointly trained with the classifier. Since the trained EBM can estimate the probability density of the input data, it can be used to detect outliers. By utilizing this property, pseudo-label learning can be performed while rejecting outliers from the unlabeled data.

\section*{Appendix}
The hybrid model used in EBPL cooperates the softmax-based classifier and EBM in the parameter space, and the third term of Eq.~(\ref{eq:loss_function}) controls the strength of the connection between the two models. The connection between the two models is formulated by considering the association of parameters between them. To align the left-hand side of Eq.~(\ref{eq:EBM}) with that of Eq.~(\ref{eq:Softmax}), we calculate the class posterior probabilities using Eq.~(\ref{eq:EBM}) and Bayes' theorem as follows:
\begin{flalign}
\label{eq:PosteriorByGauss}
p(c \!\mid\! \bm{x},\! \bm{\theta}) &\!=\! \frac{p(\bm{x}, c \!\mid\! \bm{\theta})}{\sum\limits_{c'=1}^C p(\bm{x}, c' \!\mid\! \widetilde{\bm{\theta}})} \nonumber \\
&\!=\! \frac{\exp{\left[\bm{\mu}_{c}^\top{\bf \Sigma}^{-1}\bm{x} \!+\! \ln \pi_c -\frac{1}{2}\bm{\mu}_{c}^\top{\bf \Sigma}^{-1}\bm{\mu}_{c}\right]}}{\sum\limits_{c'=1}^C \!\exp{\left[\bm{\mu}_{c'}^\top{\bf \Sigma}^{-1}\bm{x} \!+\! \ln \pi_{c'} \!-\!\frac{1}{2}\bm{\mu}_{c'}^\top{\bf \Sigma}^{-1}\bm{\mu}_{c'}\right]}}. 
\end{flalign}
The second-order terms and constant terms are canceled out because the covariance matrix is common to all the classes.\par

Comparing Eqs.~(\ref{eq:Softmax}) and (\ref{eq:PosteriorByGauss}) indicates that they have a similar formulation although the parameters are different. Specifically, there is an association between $\bm{w}_{c}$ and $\bm{{\bf \Sigma}^{-1}\mu}_{c}$, as well as between $b_c$ and $\ln \pi_c -\frac{1}{2}\bm{\mu}_{c}^\top{\bf \Sigma}^{-1}\bm{\mu}_{c}$. In model learning, the classifier and EBM work in cooperation by applying an $L_2$ loss between these associated parameters. 

\bibliographystyle{IEEEbib}
\bibliography{strings,refs}

\end{document}